\documentclass[10pt,conference]{ieeeconf} 

\IEEEoverridecommandlockouts % This command is only needed if you want to use the \thanks command
\usepackage[utf8]{inputenc}
\usepackage{graphicx}
\usepackage{amsfonts}
\usepackage{amsmath}
\usepackage{amssymb}
\usepackage{tabularx}

\usepackage [english]{babel}
\usepackage [autostyle, english = american]{csquotes}
\MakeOuterQuote{"}

\usepackage{algorithm}
\usepackage{algpseudocode}% http://ctan.org/pkg/algorithmicx
\usepackage{xcolor}
\usepackage{placeins}

\usepackage{hyperref}
\hypersetup{
    colorlinks=true,
    linkcolor=blue,
    filecolor=magenta,      
    urlcolor=cyan,
    pdftitle={Overleaf Example},
    pdfpagemode=FullScreen,
    }

\def\vx{\boldsymbol{x}} % vector x
\def\vv{\boldsymbol{v}} % vector v
\usepackage[font=footnotesize]{caption}
\usepackage{subcaption}

\def\setX{\mathcal{X}}

\def\setV{\mathcal{V}}
\def\setE{\mathcal{E}}
\def\gve{\mathcal{G(\setV,\setE)}}

\usepackage[capitalize]{cleveref}

\title{\LARGE \bf Experiments in Adaptive Replanning for \\Fast Autonomous Flight in Forests}
\author{Laura Jarin-Lipschitz, Xu Liu, Yuezhan Tao, and Vijay Kumar
\thanks{We gratefully acknowledge the support from ARL Grant DCIST CRA W911NF-17-2-0181, IoT4Ag Engineering Research Center via NSF Cooperative Agreement Number EEC-1941529, and C-BRIC, a Semiconductor Research Corporation Joint University Microelectronics Program cosponsored by DARPA. The first author acknowledges support from the NSF Graduate Research Fellowship Program under Grant No. DGE-1845298. The authors also acknowledge Avraham Cohen, Alex Zhou, and Fernando Cladera for their work on the hardware platform.}
\thanks{The authors are with the GRASP Laboratory, University of Pennsylvania, PA, 19104, USA {\tt\footnotesize \{laurajar, liuxu, yztao,  kumar\}@seas.upenn.edu.}}}

\begin{document}

\maketitle
\begin{abstract}
Fast, autonomous flight in unstructured, cluttered environments such as forests is challenging because it requires the robot to compute new plans in realtime on a computationally-constrained platform. In this paper, we enable this capability with a search-based planning framework that adapts sampling density in realtime to find dynamically-feasible plans while remaining computationally tractable.
A paramount challenge in search-based planning is that dense obstacles both necessitate large graphs (to guarantee completeness) and reduce the efficiency of graph search (as heuristics become less accurate).
To address this, we develop a planning framework with two parts: one that maximizes planner completeness for a given graph size, and a second that dynamically maximizes graph size subject to computational constraints.
This framework is enabled by motion planning graphs that are defined by a single parameter---\textit{dispersion}---which quantifies the maximum trajectory cost to reach an arbitrary state from the graph.
We show through real and simulated experiments how the dispersion can be adapted to different environments in realtime, allowing operation in environments with varying density. The simulated experiment demonstrates improved performance over a baseline search-based planning algorithm. We also demonstrate flight speeds of up to 2.5m/s in real-world cluttered pine forests.
%This parameter also drives the generation of motion primitives for the planner. The more the clutter, the smaller this parameter must be, and the finer the discretization of the input space, leading to a denser graph with an increased computational complexity. For sparser environments, a higher dispersion corresponds to a coarser discretization of the input space, leading to faster calculations.  
\end{abstract}

\section{Introduction}

Autonomous flight relies on quickly computing optimal motion plans. Recently, autonomous UAV systems have been demonstrated in GPS-denied, moderately cluttered environments with lightweight sensors~\cite{Oleynikova2020} and fast flight capabilities \cite{Mohta2018a}. However, these approaches lack the ability to fly fast when moving through environments with variable or dense clutter. Having the ability to adapt to the environment density is key to long-range efficient and reliable autonomy, especially considering the onboard computation constraints and realtime operation requirements.

Most robotic motion planning algorithms can be classified as optimization-based or sampling-based. Optimization-based planners mainly rely on the construction of 'safe flight corridors' \cite{Liu2017a} or signed distance fields/gradients \cite{Gao2017} \cite{Zhou2021} to provide constraints to an optimization problem. These planners can suffer from local minima that lead to planner failure or preclude finding highly dynamic feasible plans. On the contrary, sampling-based algorithms form plans that go through discrete vertices sampled from the underlying continuous state space, which can enable completeness guarantees by design \cite{Jarin-Lipschitz2021}. These vertices can be randomly sampled \cite{Lavalle1998} or, alternatively, chosen via searching through a set of motion primitives \cite{Pivtoraiko2009} \cite{Liu2017} (\textit{search-based planning}). Random sampling planners can only achieve probabilistic completeness guarantees, which can be a problem for realtime computational needs of robots, in contrast to the search-based planners.  The main challenge for search-based planners is that they can suffer computationally from having to evaluate many samples, especially in high dimensional state spaces. Our planner improves the computational performance of search-based planning with motion primitives in order to quickly find optimal flight plans in cluttered environments.

\begin{figure}[!t]
	\centering
	\includegraphics[ width=.99\columnwidth]{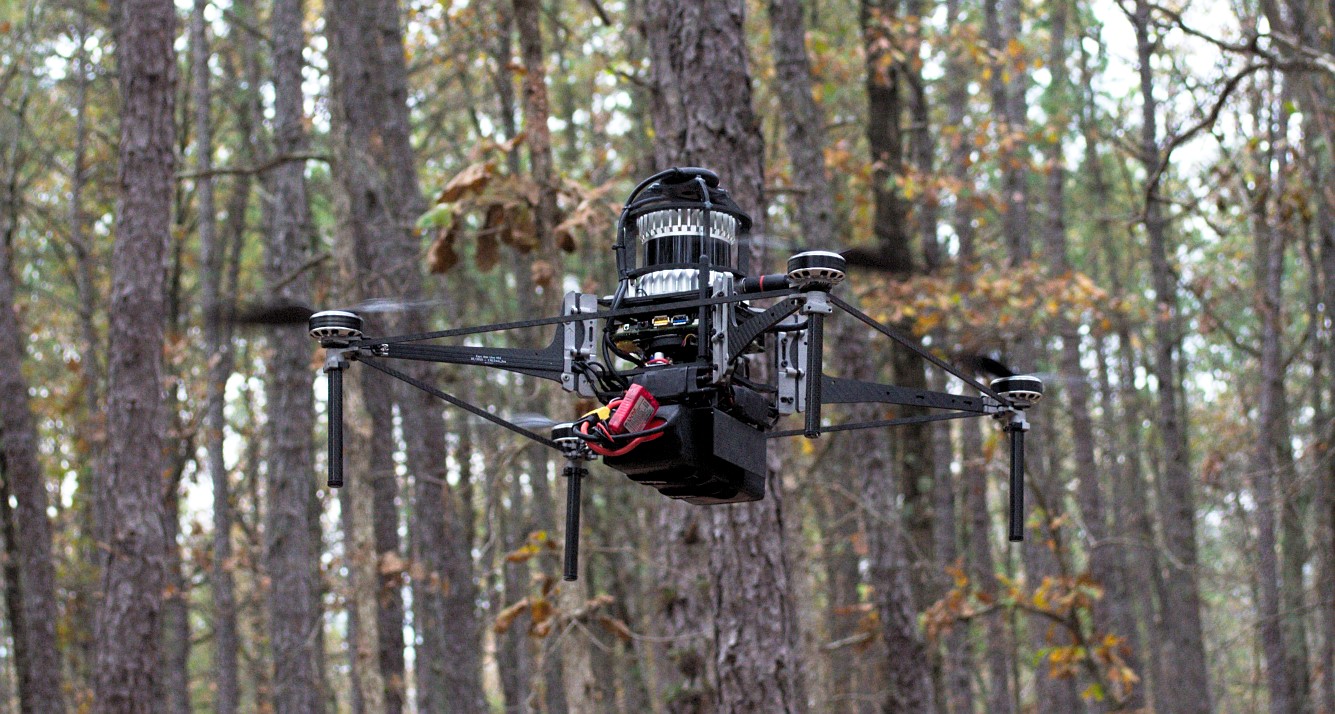}
	\caption{We performed our experiments on our customized "Falcon 4" UAV platform, as described in \cite{Liu2021}. This platform carries an IMU, stereo cameras and a 3D LIDAR. The onboard Intel NUC 10 computer is responsible for running the navigation stack and storing data.}
	\label{fig:quad}
	        \vspace{-0.2in}
\end{figure}

Search-based planning is computed with regular graph search algorithms such as Dijkstra's or A*. Therefore, computational performance degrades as the number of vertices and edges in the graph increases. Sampling cluttered environments in search-based planning with motion primitives is made especially difficult due to two computational challenges: first, if there are narrow corridors in the environment, a large and therefore expensive graph will be needed to find a plan; second, performing graph search is inherently less efficient in cluttered environments due to heuristics more severely underestimating the cost-to-go. Despite how expensive larger graphs make computation, most search-based planning algorithms do not deliberately select their vertices to be placed optimally with respect to completeness, and instead choose them in ways that are easy to compute such as uniform input sampling \cite{Liu2017}. Our prior work \cite{Jarin-Lipschitz2021} began to address this problem by designing search-based planning graphs that ensured that any state in the state space was optimally nearby to a vertex in the graph, which is encoded by a single parameter, dispersion. However, it still left open the question of how large of a graph (how small of a dispersion) to use when planning.

Accordingly, we develop a search-based planner that is able to \textit{adapt} to the environment by changing the dispersion of the motion planning graph used.
The contributions of our work are as follows:
\begin{enumerate}
    \item An adaptive, minimum dispersion search-based motion planning algorithm.
    \item Simulated results comparing our method with the state-of-the-art in search-based planning for UAVs, at speeds up to 7 m/s in cluttered forest environments.
    \item Real-world experimental results in a cluttered forest environment with varying densities of obstacles, at speeds up to 2.5 m/s.
\end{enumerate}

\section{Related Work}

Our planner lies in the category of search-based motion planning. Search-based motion planning, also sometimes referred to as lattice planing, leverages the widely studied field of graph traversal algorithms in order to plan robot trajectories. Specifically, our work, a continuation and experimental implementation of \cite{Jarin-Lipschitz2021}, focuses on optimizing the planning graph's design.

Preliminary search-based planners' graph design utilized uniform discretization in the state space \cite{Pivtoraiko2009}. Later work was able to achieve realtime performance onboard a UAV platform by uniformly \cite{Liu2017} or randomly \cite{Dharmadhikari2020} sampling primitives in the \textit{input} space (jerk and acceleration respectively) over a constant time, which severely limits the dynamic range of robot maneuvers. These sampling schemes are largely arbitrary, and do not ensure a minimal graph design, which we have noted is necessary for optimal computational performance of search-based motion planning.

These search-based motion planners can be extremely sensitive to their resolution or parameterizations (e.g., discretization of the control space and the time horizon for the motion primitive) and may require tuning to each novel environment, since certain discretizations will lead to planner completeness while others will not. Therefore, such planners usually require a good prior understanding of the environment and an experienced expert to carefully choose parameters. This severely limits the system's capability to operate in large-scale environments where the distribution of obstacles are previously unknown or hard to model. By contrast, our prior work in \cite{Jarin-Lipschitz2021} has only a single parameter that increases planner completeness and optimality, as shown in \cref{fig:dispersion_vs}.

Other prior work (some not feasible in realtime) has noted the utility of multi-resolution sampling-based planning strategies such as \cite{Likhachev} \cite{Du2016} \cite{Pivtoraiko2013} \cite{Cohen2011}, since we may not know whether we will be in a difficult (cluttered) or easy (free) environment at plan time. However, these approaches all rely on some type of manual or uniform sampling in the state space in the \textit{Euclidean metric}, which is not meaningful with respect to comparing two robot states. For our planner, we instead adapted the minimum dispersion framework of \cite{Palmieri2019} in order to optimally sample the state space with respect to the \textit{trajectory cost metric} which defines planning optimality.

\section{The Experimental System}
\subsection{Unmanned Aerial Vehicle}
Our autonomous UAV system, pictured in \cref{fig:quad}, has a similar design as the one presented in our recent work \cite{Liu2021}. The system consists of three main modules: state estimation and mapping, planning, and control, and is equipped with a Ouster 3-D LiDAR, a stereo camera synced to a vectornav VN100 IMU on a custom OVC3 computer~\cite{Quigley2019}, an Intel NUC 10 computer and a 17,000mAh lithium-ion battery capable of 30 minute flight times. We use S-MSCKF stereo VIO algorithm~\cite{Sun2018} which provides with with 30Hz robust odometry. Similar to the state estimation module proposed in our prior work~\cite{Mohta2018a}, the VIO is then feed into an Unscented Kalman Filter (UKF) which runs at the same rate as the IMU (200Hz). The output of the UKF is used for the control loop. The global and local voxel maps are built simultaneously based on the 3D point cloud data from LIDAR and the robot pose. 

\begin{figure*}[t!]
        \centering
        \begin{subfigure}[t]{.99\columnwidth}
        	\centering
        	\includegraphics[trim=0cm 5cm 0cm 0cm, clip, height=2.5cm]{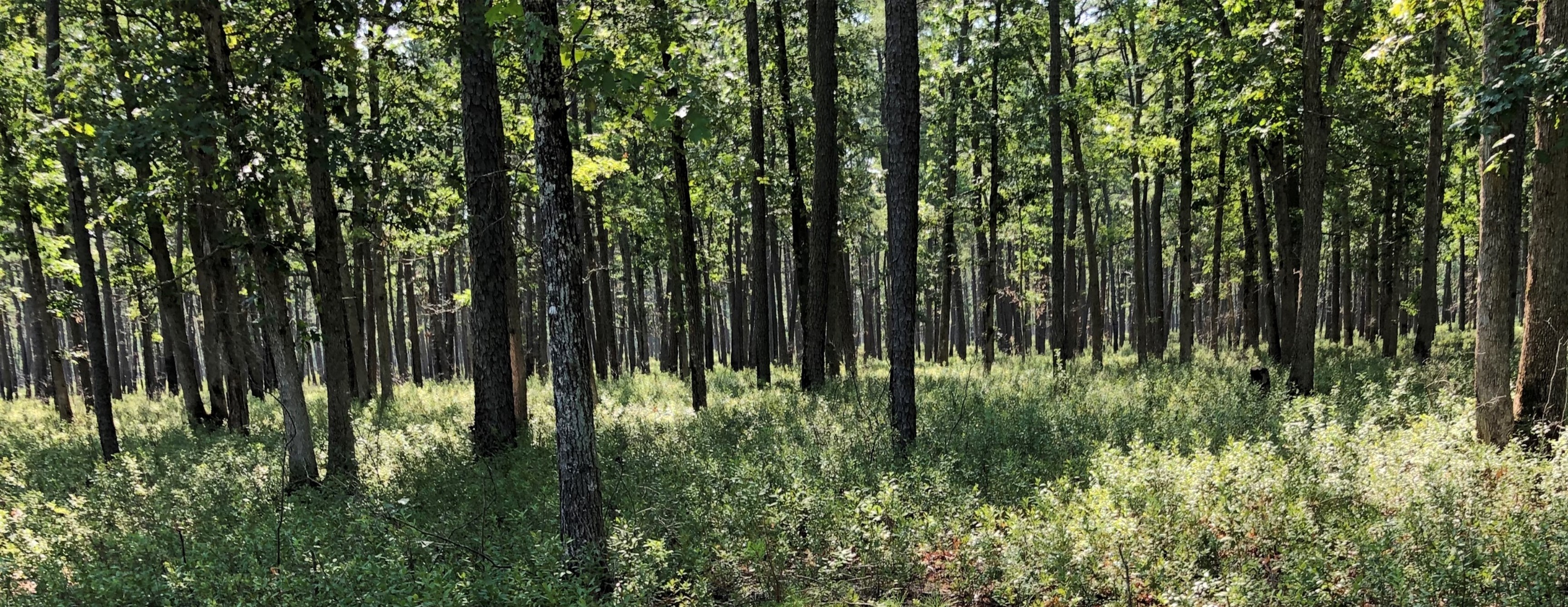}
        \end{subfigure}
        \begin{subfigure}[t]{.99\columnwidth}
            \centering
        	\centering
        	\includegraphics[trim=0cm 0cm 0cm 3cm, clip, height=2.5cm]{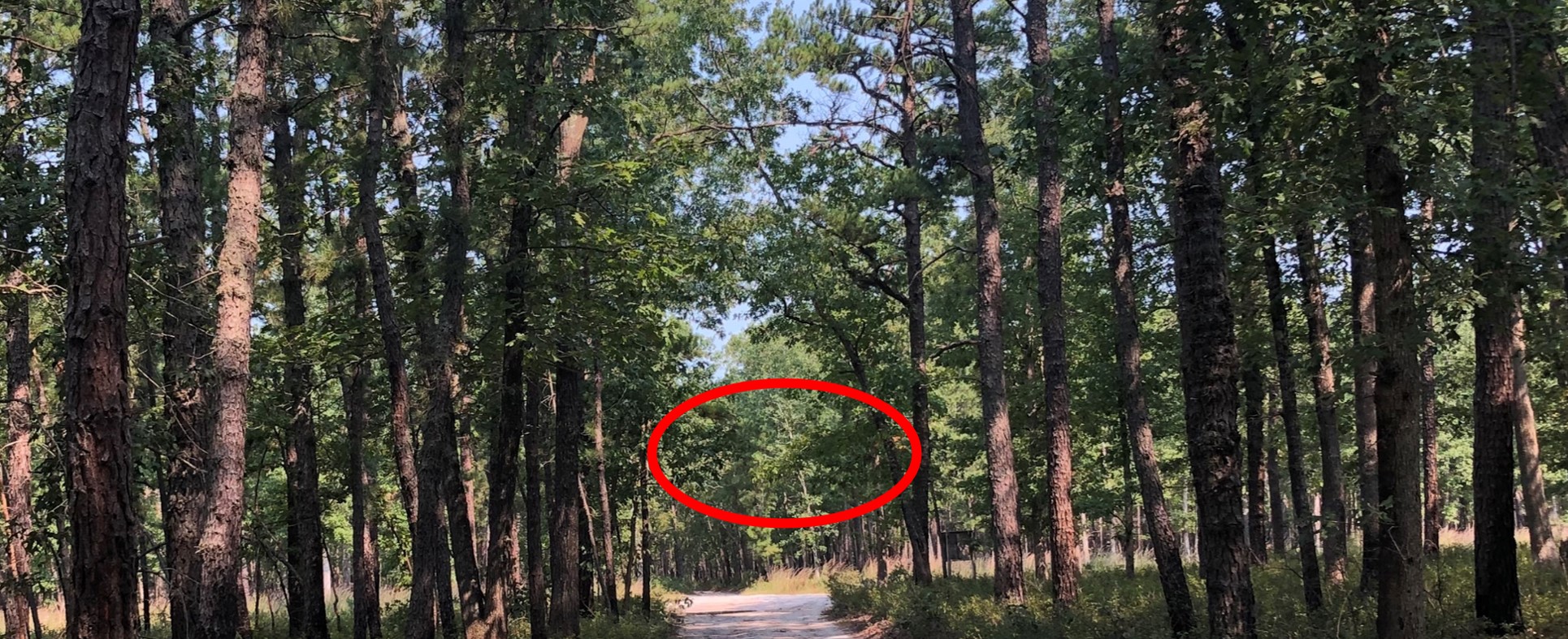}
	        \end{subfigure}
        	\caption{\textit{The left panel} shows a view of the under-canopy environment where our autonomous flight experiments were performed. \textit{The right panel} is an illustration of the varying density along the trail section of the environment. The planner's replan time spiked once when the UAV was flying towards the end of this trail. This is because that there are two trees whose canopies were closer to the ground (highlighted in red circle), forming a narrow gap. The proposed planner was able to adaptively choose the dispersion to keep the replan time within the given budget.}
    	\label{fig:illustrate-density}
        % \vspace{-0.2in}
\end{figure*}

In a concurrent work \cite{Liu2021} we illustrate realtime semantic mapping using onboard sensors and the use of this semantic information for state estimation and mapping for under-the-canopy flight in an environment similar to that in \cref{fig:illustrate-density}. This work relies on a traditional non-adaptive uniform-sampling based motion primitive planner \cite{Liu2017}. In contrast, this paper presents a new adaptive minimum-dispersion-based motion primitive planner that helps to plan more optimal trajectories as well as increase the mission success rate in forest-like environments.

\subsection{Local-Global Architecture}
We adopt the same planner and mapper framework as presented in \cite{Liu2021}, which includes a global and local map, with corresponding plans. The global (re)planner uses a jump point search algorithm to search for the collision-free shortest path to the global goal in a coarsely-discretized voxel map for better efficiency. The local (re)planner works in a finely discretized, small and robot-centric local map for better optimality and traversability. The goal given to the local planner is found from intersecting the global plan with the local map. The adaptive minimum dispersion planner described in this work functions as the local planner. (Theoretically, it could also be used in the future for the global planner at high dispersions.) The local planner runs at a higher frequency than the global planner, and the local planner is required to meet a time threshold. If it fails to find a plan in its allotted time, a stopping planner is called to ensure safety. %By using this replanner, the autonomous flight system is demonstrated to be able to adapt to the clutteredness of the environment in realtime, thus planning more successfully and optimally than it would at a fixed resolution.
\begin{figure}[t!]
	\centering
	\includegraphics[width=.85\columnwidth]{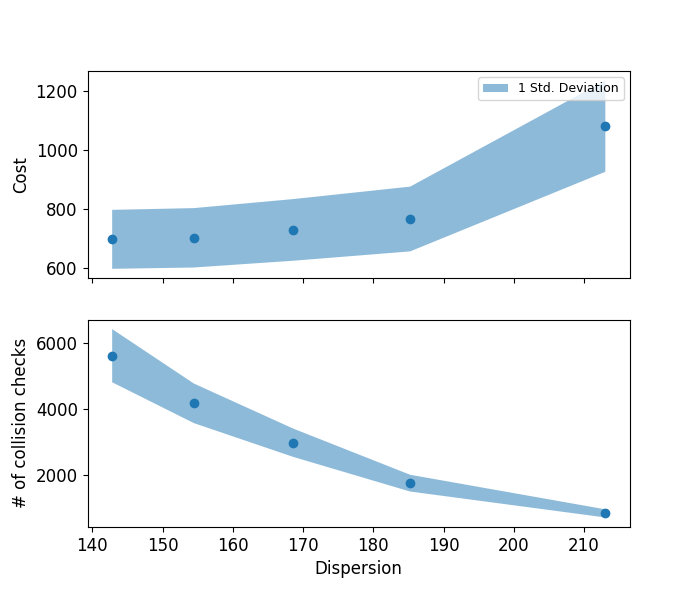}
	\includegraphics[width=.42\columnwidth]{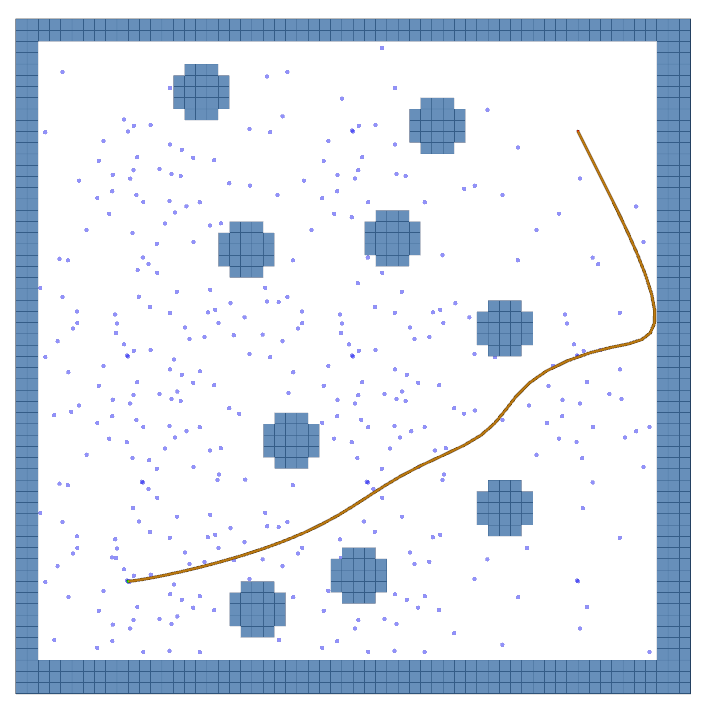}
	\includegraphics[width=.42\columnwidth]{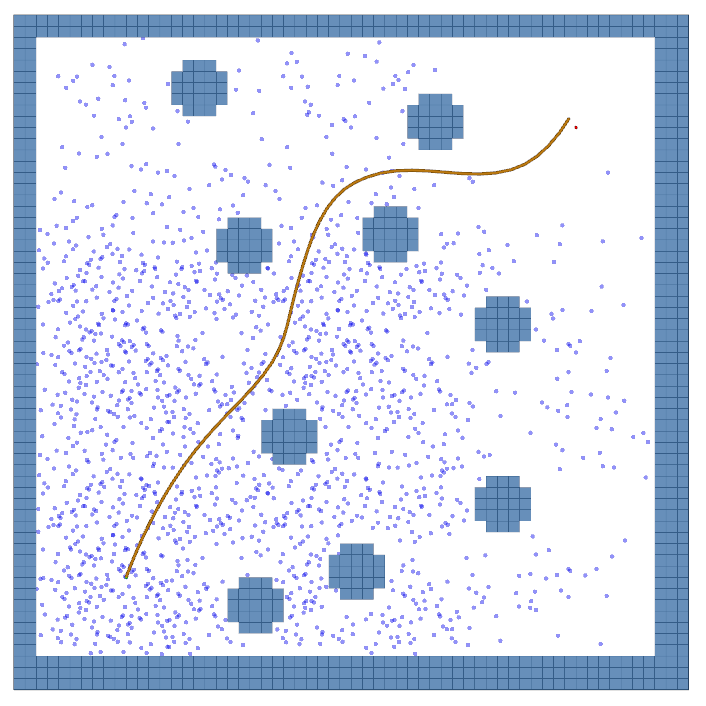}

	\caption{A useful property of the dispersion is that it allows us to trade off computation time with plan optimality. The top plots show the result of planning in 100 randomized maps with the same number, size, and spacing of obstacles, with graphs of 5 different dispersions, max speed 7 $m/s$, and max acceleration 4 $m/s^2$. We can see that lower dispersion leads to more optimal plans (top), but more computational cost (middle) using collision checks, which dominates planning time, as a proxy. For this not very cluttered environment, we reach an asymptote on optimality. The bottom figures correspond to one of the 100 planning problems, with the highest (left) and lowest (right) dispersion graphs. The blue dots represent the graph search's visited states projected into the configuration space, while the brown curve is the plan. The low dispersion graph search checks many more states, but returns a more optimal plan.}
	\label{fig:dispersion_vs}
        \vspace{-0.15in}

\end{figure}

%%%%%%%%%%%%%%%%%%%%%%%%%%%%%%%%%%%%%%%%%%%%%%%%%%%%%%%%%%%%%%%%%%%%%%%%%%%%%%%%%%%
%%%%%%%%%%%%%%%%%%%%%%%%%%%%%%%%%%%%%%%%%%%%%%%%%%%%%%%%%%%%%%%%%%%%%%%%%%%%%%%%%%%

%%%%%%%%%%%%%%%%%%%%%%%%%%%%%%%%%%%%%%%%%%%%%%%%%%%%%%%%%%%%%%%%%%%%%%%%%%%%%%%%%%%
\begin{figure*}[t!]
        \centering
        \begin{subfigure}[t]{.99\columnwidth}
        	\centering
        		        \includegraphics[trim=0cm 0cm 0cm 0cm, clip, height=4cm]{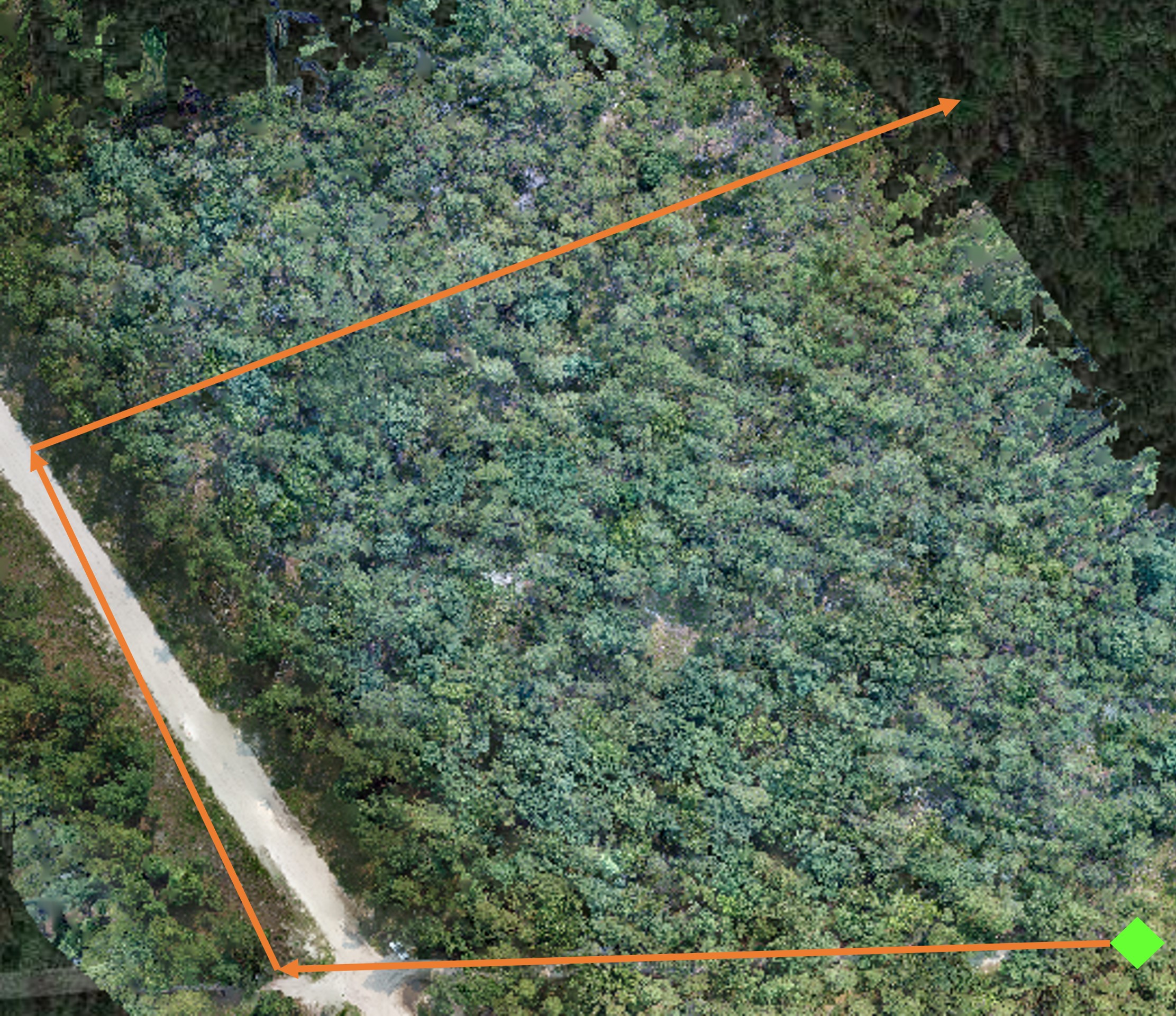}
        \end{subfigure}
        \begin{subfigure}[t]{.99\columnwidth}
            \centering
        		\includegraphics[trim=3cm 4cm 1cm 3cm, clip, height=4cm]{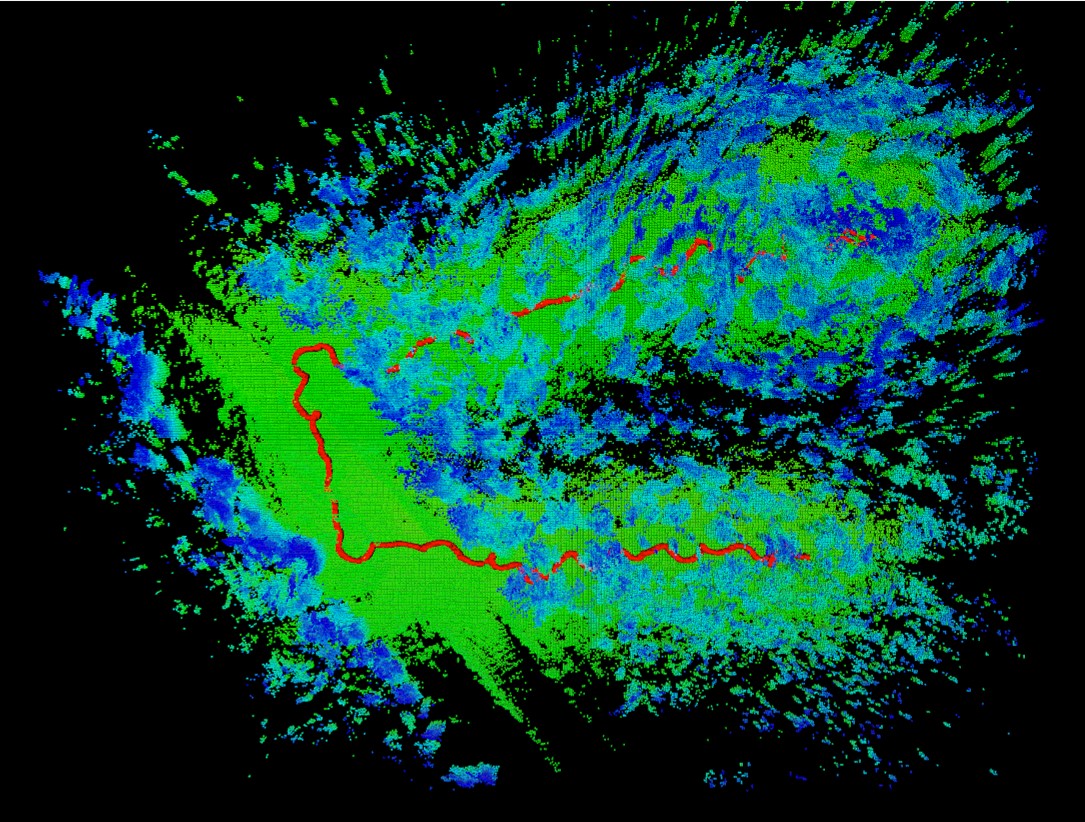}
	        \end{subfigure}
        	\caption{\textit{The left panel} shows a mission overview. The orange lines show the mission published by the user. The green diamond is the home start position. This overhead view is generated by stitching the images gathered by our over-canopy UAV. \textit{The right panel} is a top view of trajectory and the global voxel map built by the robot. The map is shown as colored points where the color encodes the height of the voxel, and the trajectory is shown in red curve. The total trajectory length is 380 meters. The clutteredness varies significantly along this mission.}
                    \label{fig:experiment-map-and-traj}
        % \vspace{-0.2in}
\end{figure*}
%%%%%%%%%%%%%%%%%%%%%%%%%%%%%%%%%%%%%%%%%%%%%%%%%%%%%%%%%%%%%%%%%%%%%%%%%%%%%%%%%%%

\section{Planning Adaptively with Minimum Dispersion Graphs}
\subsection{Search-based Motion Planning}
Given a graph made up of vertices $\setV$ and edges $\setE$, $\gve$, a planning query will return a set of dynamically feasible edges that link a start state to a goal state, through obstacle-free space. Search-based motion planning is a sampling-based motion planning technique, relying on discretization to make motion planning computationally feasible. However, there are infinite possible discretizations of the state and input spaces, many of which will not suffice to create a computationally feasible or efficient planning solution, so choosing the graph discretization in a principled manner is of paramount importance.

\subsection{Minimum Dispersion Search-based Motion Planning Graphs}
The main module of our planning algorithm is largely as described in our prior work \cite{Jarin-Lipschitz2021}. We first generate search-based motion planning graphs offline, in an effort to create optimally \textit{sparse} motion planning graphs. Since the online portion of search-based motion planning, graph search, is $O(|E| + |V|log(V))$ (for Dijkstra's), the way we can improve the performance of search-based planning is by focusing on reducing the number of edges and vertices required to find a plan. The graphs are generated in free space, since we cannot know what points in the state space will be occupied at plan time, leaving the online graph search to prune out the colliding and suboptimal trajectories.

To do this, we generate graphs using an algorithm originally proposed in \cite{Palmieri2019} that greedily minimizes \textit{dispersion} in the state space using an optimal trajectory cost metric. 

Given a vertex set $\setV$ and a cost metric between two states $J$, dispersion $d$ is defined in \cite{nied} over the state space $\setX$ as follows:
\begin{align} \label{eq:disp}
    d(\setV) = \sup\limits_{\vx \in \setX} \left[ \min_{\vv \in \setV} J(\vx, \vv)\right]
\end{align}

One of the key insights we provided in our previous work was to apply the definition of dispersion to systems with \textit{drift}, i.e. when solving a boundary value problem between two states, which one is the starting state matters because of the robot has momentum. This led us to an updated version of the dispersion for systems with drift:

\begin{align}
    d(\setV) = \sup\limits_{\vx \in \setX} \left[ \min_{\vv \in \setV} [\max(J(\vx, \vv),J(\vv,\vx))]\right]
\end{align}

By changing the dispersion definition, we were then able to prove for real robotic systems with drift, that given that a true optimal plan exists from start to goal and it is at least a clearance $\delta$ away from any obstacles in the trajectory cost metric, we are guaranteed to find a plan through our search-based planning graph if the dispersion of the graph is less than $\delta/2$. However, in practice, it is impossible to know what the clearance of the true optimal plan is. This leads us to the adaptive strategy we will describe in the following sections. Though we do not know what the clearance of the true optimal trajectory is, we know that if we decrease the dispersion, we can only get more complete.

It was later found empirically that the following slightly different definition of the dispersion maintained the optimality vs. computation time properties described later in \cref{fig:dispersion_vs} and performed better computationally, but is presented as yet without a proof of completeness. 

\begin{align}
    % d(\setV) = \sup\limits_{\vx \in \setX} \left[ \min_{\vv \in \setV} [\max(J(\vx, \vv),J(\vv,\vx))]\right]
    d(\setV) = \sup\limits_{\vx \in \setX} \left[ \max\left( \min_{\vv_1 \in \setV} [J(\vv_1, \vx)], \min_{\vv_2 \in \setV} [J(\vx, \vv_2)]\right) \right]
\end{align}

Conceptually, in the first definition, the dispersion is determined by the worst-case difficulty of travelling back and forth between a state and a nearby vertex. In the second definition, we instead consider the difficulty of traveling back and forth between a state and the graph itself, such that the "source" and "destination" vertices need not be the same.

\subsubsection{Boundary Value Problem}
As described in \cite{Jarin-Lipschitz2021}, in order to generate the minimum dispersion planning graph,  it is required to compute boundary value problems between pairs of states, also called motion primitives. In \cite{Jarin-Lipschitz2021}, the motion primitives were computed with a nonlinear trajectory optimization that minimized a weighted sum of trajectory time and the norm of the jerk, under velocity and acceleration limits. However, this optimization problem was very slow (.5-1s), leading to the graph generation, in which thousands of boundary value problems must be computed, to be extremely slow. Additionally, this makes solving planning queries from arbitrary start states that are not on the pre-generated graph prohibitively expensive, since new boundary value problems must be solved online in order to start the graph search. In this work, we updated our graph generation algorithm to use the nonlinear trajectory optimization from \cite{Burri2015}\cite{Richter2016}, which has a similar cost function but far superior computational performance. It provides plans in the bounded state space of (position, velocity, acceleration) while minimizing jerk and time in the cost function.

Although we use the same trajectory optimization as \cite{Burri2015}, it is important to note the large differences in our approach, owing to our method being search-based while theirs is optimization-based. While they use an RRT* planner to find straight line paths which are then fed to the trajectory optimization, our work creates graphs of dynamically feasible polynomial segments, which are combined together in novel arrangements at graph search time. This allows our planner to produce much more dynamic plans, since we are not constrained to straight line paths that may be highly suboptimal when e.g. starting velocity is considered.

\subsubsection{Relationship of Dispersion to Plan Optimality and Computation Time}
\cref{fig:dispersion_vs} shows an empirical comparison of dispersion with trajectory cost and computational cost. It also shows an example of two plans at differing dispersions. We use this as evidence and motivation for the ability to plan with variable dispersion, depending on computational constraints. Compared to other state of the art search-based motion planners such as \cite{Liu2017}, the fact that we have only a single parameter to tune, which can be almost trivially be adapted to operating conditions, is a key advantage.

\subsection{Adaptive Planner}
One way in which our robot software system design promotes planning safety is by giving the planner an empirically determined computation time budget. If the local plan is too old, it is less safe to continue to follow it, since we may e.g. have seen new obstacles or updated our state estimate. This danger therefore scales with the speed of the flight and the clutteredness of the environment.  Some preliminary effort has been made in work such as \cite{Falanga2019} to study the theoretical interplay between some of these factors (mostly focused on perceptual latency), but more analysis is needed.

\cref{fig:dispersion_vs} provides empirical evidence that our planner can trade off plan optimality and computation time by tuning a single parameter, the graph dispersion.  \cite{Jarin-Lipschitz2021} also presents the theoretical result that the lower the dispersion is, the more complete the planner is with respect to being able to solve more difficult (lower clearance) plans. This leads us to the concept of the adaptive planner, which changes dispersion online in response to feedback from the planner. By providing the robot with not just a single planning graph, but a family of graphs that are either less computationally expensive to plan in or more optimal and complete, we can create a planner that works in a wider range of environments under computational constraints, while still providing the most optimal plan that we can afford to compute. In this way, not only is each planning graph optimally sparse, but we use the planning graph that maximizes our computation budget. 
% TODO  add cite science robotics for theoretical limit

The adaptive planner has three simple triggers to change the dispersion.
\begin{enumerate}
    \item \textit{Planner violates time constraint} - If the planning time exceeds the timeout threshold, we increase the dispersion, since a higher dispersion graph has fewer vertices and edges, and is therefore faster on average if a plan exists.
    \item \textit{Planner fails to find a plan} - If the planner exhausts the possible edges to expand and finds all trees terminating in a colliding edge, we decrease the dispersion, in order to provide the planner with more vertices that may land in a e.g. a narrow corridor that a higher dispersion graph missed.
    \item \textit{Planner is performing well with large time margin} - If the planner is not failing and is well-below the timeout threshold, we decrease the dispersion, since we have extra computational margin to compute more optimal plans.
\end{enumerate}

\begin{figure}[t!]
        % \vspace{-0.35in}
	\centering
	\includegraphics[trim=0cm 0cm 0cm 1.5cm, clip, width=.8\columnwidth]{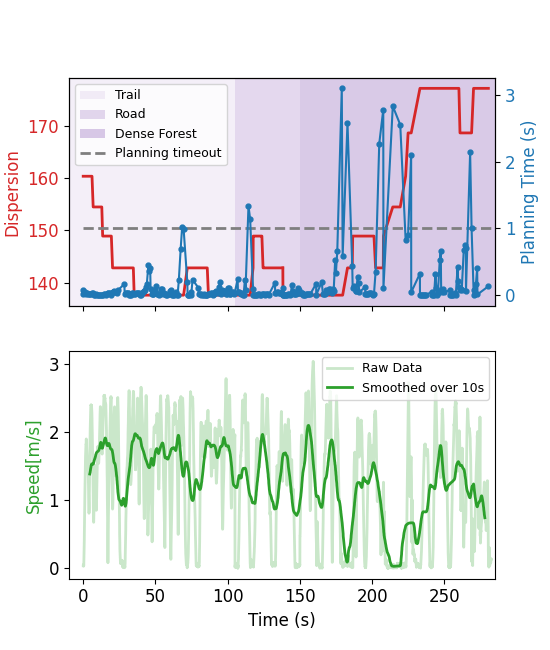}
	\caption{ This figure displays the main functionality of the adaptive planner. The planner is following the mission displayed in \cref{fig:experiment-map-and-traj}. From time 0-40s, we can see that the planner is running well beneath the timeout, so under adaptive planning trigger 3, it decides to decrease dispersion. There is one timeout at ~70s, so it momentarily increases dispersion. There are planner timeouts at ~80s (detailed more in \cref{fig:illustrate-density}) and ~110s that cause adaptive planner trigger 1, to increase dispersion to reduce planning time. From time 110-150s the UAV flies easily in the uncluttered road, so dispersion decreases again. At time ~160s, the robot reenters the forest and starts experiencing timeouts, so it increases dispersion eventually to the maximum, and is able to get deeper into the forest. %At the end, it reaches an area of the forest that is so cluttered, that the planner fails to find a plan (causing a decrease at time ~260s) with the high dispersion it requires due to the timeout constraint, so we end the mission.
	        \vspace{-0.1in}
}
	\label{fig:adaptive_planner}
\end{figure}

All three of the adaptive triggers are demonstrated in the experimental flight in \cref{fig:adaptive_planner}.

A consequence of this architecture is that the adaptive planner can also detect when the environment is too difficult to plan in with the given computational and state constraints. This can happen due to the dual computational challenges of cluttered environments mentioned earlier, that the computation cost scales with the clutteredness of the environment (which could be helped by higher dispersion), and the fact that the planner is less complete, since fewer samples land in unoccupied states (which could be helped by lower dispersion). If triggers 1 and 2 are both occurring, then there is likely no dispersion that will yield successful planning. A future extension could include reducing the state constraints (max velocity and acceleration) in order to constrict the size of the state space and therefore artificially decrease dispersion. This has the added benefit that flying slower could allow for higher planner computational budget.

\section{Results and Analysis}

To verify and characterize the proposed adaptive planner, we performed multiple experiments in both simulation and real-world forests.

\subsection{Simulated Experiments and Results}
We use a photo realistic simulator that is based on Unity 3D and compatible with ROS. The simulator provides high-fidelity imagery and LIDAR data, which can be directly used for our VIO and LIDAR state estimation and mapping system. The simulation environments are customized based on the real-world tree position data we obtained~\cite{Liu2021}. Forests that have large variance in density are generated to demonstrate the planner's capability to adaptively choose the dispersion according to the changes in environment density.

\begin{figure}[h!]
        \vspace{-0.2in}

	\centering
	\includegraphics[width=.99\columnwidth]{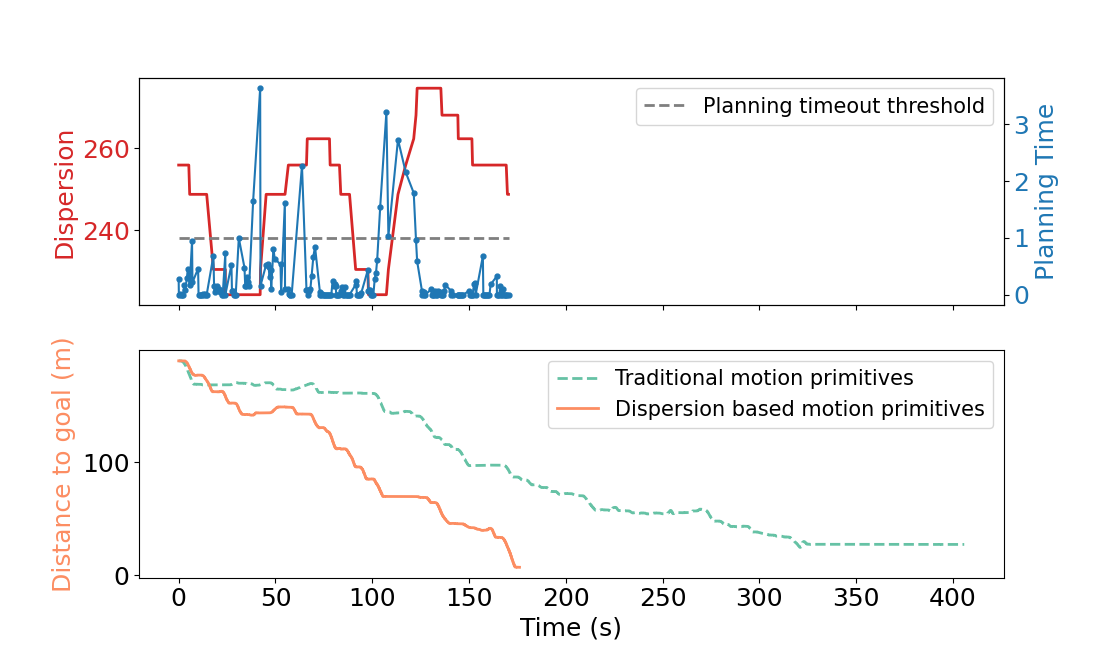}
	\caption{A simulation experiment comparing the proposed planner against a traditional uniform input sampling motion primitive planner~\cite{Liu2017}. The top panel shows the change in planning time (blue) and dispersion (red) chosen by the planner. The bottom panel shows a comparison between our planner (orange) and the uniform input (green) sampling planner in distance to goal. Our planner approaches the goal much faster, and finally reaches the goal. The benchmark planner approaches the goal slowly and finally fails to find a trajectory that reaches goal, thus failing the mission.}
	\label{fig:simulator-result}
	        \vspace{-0.1in}

\end{figure}

One experimental result is shown in \cref{fig:simulator-result}. In this experiment, we compared our planner's performance against a traditional uniform input sampling motion primitive planner~\cite{Liu2017}. Parameters such as the maximum velocity, planning horizon, replan rate, replan time budget, etc., are all set to be the same. The mission sent to the UAV is also exactly the same. The UAV takes off from a relatively sparse region of the forest (average between-tree corridor is 8.13m), and flies for a short distance before heading into a dense region of the forest (average between-tree corridor is 1.91m). The other parameters specific only to ~\cite{Liu2017} were ones that were tuned for our work in ~\cite{Liu2021}. The proposed planner was able to adapt to the change in density as shown in the top panel of \cref{fig:simulator-result}. On the contrary, the benchmark planner is unable to adapt to the environment, thus resulting in a very long time to approach the goal. It finally becomes stuck due to the density around the goal being too high, never reaching the goal.

\subsection{Real-world Experiments and Results}
Our real-world field experiments were performed at the Wharton State Forest, Hammonton, NJ, USA. This forest has 122,880 acres of the Pinelands. This forest has a large variance in tree density, shape and sizes. To verify our system's ability to adapt to the change in clutteredness of the environment, we executed several missions in the part of forest where the tree density varies significantly. We used graphs in two dimensional configuration space, as opposed to 3D, due to the computational limits of the platform.

\begin{figure}[h!]
	\centering
	\includegraphics[width=.5\columnwidth]{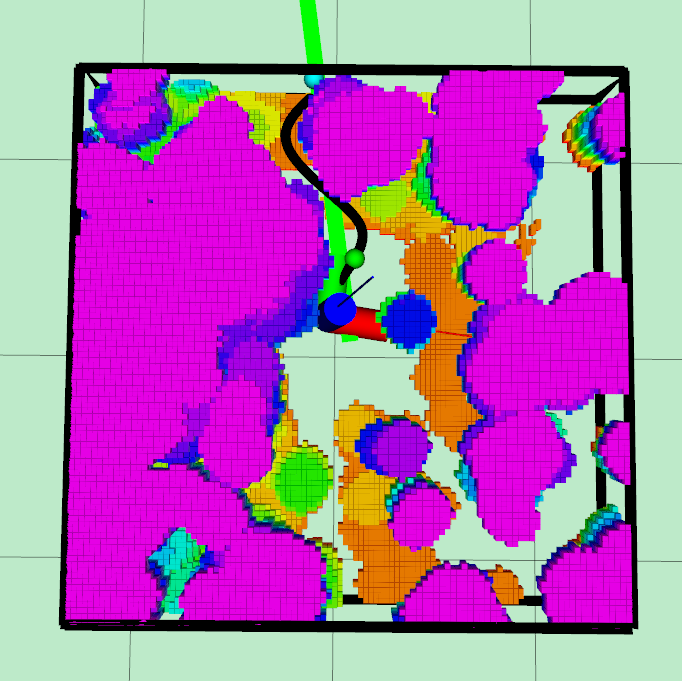}
	\caption{A snapshot of a single local plan (black) from the real-world forest experiment, on the 'trail' section. The blue sphere is the local planning goal, found from the intersection of the global plan with the local map, and the green sphere is the local planner start, slightly in the future to provide continuity with the last planned trajectory. The planner is able to find a dynamically feasible path through a narrow corridor in the inflated voxel map (rainbow).}
	\label{fig:plan_closeup}
	       % \vspace{-0.2in}
\end{figure}

One of our missions is shown in \cref{fig:experiment-map-and-traj}. This mission consists of the following phases: In phase 1, the UAV takes off from a trail, and flies along the trail to the main road to reach the 1st waypoint. In phase 2, the UAV flies along the main road to reach the second waypoint. In phase 3, the UAV flies directly into the dense under-canopy environment, and navigates in between trees with tiny branches and above thick undergrowth as shown in \cref{fig:illustrate-density}. The density of the environment during this mission changes from moderately cluttered to sparse to highly cluttered. The total trajectory length is 380 m. As shown from the \cref{fig:adaptive_planner}, the planner is able to adaptively choose the dispersion according to the environment clutteredness. 

During phase 1, the planner was performing well in the beginning (first 70s in \cref{fig:adaptive_planner}), with the average replan time well below the planning time budget, thus increasing decreasing the dispersion to gain more optimality. A snapshot of one motion plan during this time is displayed in \cref{fig:plan_closeup}. At the end of phase 1, however, the replan computation time reached exceeds the given budget. This was caused by the sudden increase of clutteredness in the end of the trail, as illustrated in \cref{fig:illustrate-density}. The planner was able to adapt to this situation rapidly and increased the dispersion, which led to the decrease of replan time. During phase 2, since the UAV was flying on the main road, the replan time was constantly low, except spiking for once (at $\sim$120s). That was because the robot was surrounded by the tall undergrowth on the roadside, which lies in the lower left corner of \cref{fig:experiment-map-and-traj}. During phase 3, the planner started to exceed the given time budget as soon as the UAV entered the under-canopy environment as shown in \cref{fig:illustrate-density}. The adaptive strategy was able to adjust the dispersion, so that the UAV kept flying through the dense trees. When the dispersion reached high enough value, the replan time dropped back and stayed within the given budget for the most time. 

The UAV was flying reliably during the whole flight, even though the density of the environment varied significantly. This demonstrates the superiority of our planner to a traditional non-adaptive planner. If the baseline planner encountered the same timeout situation due to the increase in environment density, it would have a high chance of repeatedly timing out and failing to complete the mission, as was shown in the simulated results. 

Though our experimental results are presented in a forest in order to challenge the adaptiveness of the planner, the proposed planner is not restricted to forests. So far, we have also demonstrated its performance in other settings such as an outdoor campus environment, at speeds of up to 5.3m/s (as shown in the accompanying video).

\section{Conclusion}
In summary, we presented an adaptive, minimum-dispersion search-based motion planning algorithm for operation in both uncluttered and cluttered environments. The framework relies on the choice of a single parameter, the dispersion, which is adapted to the environment. A sparse environment leads to a lower dispersion (due to the decreased computational cost of planning) which in turn corresponds to motion primitives with more aggressive, faster flight. Conversely, the dispersion is increased as the environment is more cluttered leading to relatively slower flights that are feasible to compute in the planning time budget.  We validated our approach on real and simulated forest environments with variable clutter, with  speeds up to 2.5 m/s in pine forests. Additional simulation and benchmarking results will accompany the forthcoming open source release of the motion planning code base at \url{https://github.com/ljarin/dispersion_motion_planning}.

The same approach can be used to compute 3D plans which will allow us to increase our mission success in complex environments. Another practical challenge we hope to overcome is that due to the quasirandomness that underlies the dense state sampling set from which our vertices are chosen, the robot rarely flies perfectly straight when in a completely uncluttered environment, instead following an 'S' path. 

On the theoretical side, we are hoping to show that the minimum dispersion graph is within a constant factor of being as sparse as possible for search-based motion planning for a given dispersion, and that the updated dispersion definition still holds completeness properties.

More broadly, we hope to integrate the semantic mapping capability from our concurrent work in order to improve localization and obstacle detection and therefore improve planner performance, allowing us to fly faster and safer.

\bibliographystyle{ieeetr}
\bibliography{_dissertation}

\end{document}